\documentclass{article}
\usepackage{authblk}
\usepackage[utf8]{inputenc}
\usepackage{diagbox}
\usepackage{amsmath}
\usepackage{amsfonts}
\usepackage{graphicx}
\usepackage{mathtools}
\usepackage{float} 
\usepackage{xcolor}
\usepackage{hyperref}
\usepackage{verbatim}
\usepackage{float}

\title{Physics-Informed Neural Network based inverse framework for time-fractional differential equations for rheology}
\date{}
\author[1]{Sukirt Thakur}

\affil[1]{\small{School of Mechanical Engineering,
            Purdue University, 
            West Lafayette,
            47907, 
            Indiana,
            USA}}

\author[1]{Harsa Mitra}
\author[1]{Arezoo M. Ardekani}

\begin{document}

\maketitle

\begin{abstract}
Time-fractional differential equations offer a robust framework for capturing intricate phenomena characterized by memory effects, particularly in fields like biotransport and rheology. However, solving inverse problems involving fractional derivatives presents notable challenges, including issues related to stability and uniqueness. While Physics-Informed Neural Networks (PINNs) have emerged as effective tools for solving inverse problems, most existing PINN frameworks primarily focus on integer-ordered derivatives.

In this study, we extend the application of PINNs to address inverse problems involving time-fractional derivatives, specifically targeting two problems: 1) anomalous diffusion and 2) fractional viscoelastic constitutive equation. Leveraging both numerically generated datasets and experimental data, we calibrate the concentration-dependent generalized diffusion coefficient and parameters for the fractional Maxwell model. We devise a tailored residual loss function that scales with the standard deviation of observed data.

We rigorously test our framework's efficacy in handling anomalous diffusion. Even after introducing 25\% Gaussian noise to the concentration dataset, our framework demonstrates remarkable robustness. Notably, the relative error in predicting the generalized diffusion coefficient and the order of the fractional derivative is less than 10\% for all cases, underscoring the resilience and accuracy of our approach. In another test case, we predict relaxation moduli for three pig tissue samples, consistently achieving relative errors below 10\%. Furthermore, our framework exhibits promise in modeling anomalous diffusion and non-linear fractional viscoelasticity.  
\end{abstract}
\section{Introduction}
Fractional calculus broadens the traditional concepts of differentiation and integration to include non-integer orders. Its roots trace back to 1695, when Leibniz posed the question of defining the $\frac{1}{2}$th order derivative in a letter to L'Hospital. Since then, numerous mathematicians have contributed to its evolution, shaping its theoretical foundations. While fractional calculus is often perceived as distinct from integer-order calculus, it essentially serves as an extension, offering a broader and more natural framework. Here, integer-order derivatives emerge as special cases within this framework \cite{Podlunby1998}.

The practical relevance of fractional calculus has significantly increased since the 1970s, when it gained attention from applied scientists and engineers. This surge in interest stemmed from the recognition that fractional calculus provides enhanced models for many intricate phenomena across various domains. These include electromagnetism \cite{Engheta1996}, rheology \cite{Mainardi2010, Bonfanti2020}, fluid mechanics \cite{Odibat2009}, and biology \cite{Anna2018}. Fractional calculus excels in capturing the memory and non-local effects inherent in these systems, aspects often inadequately described by integer-order derivatives. Fractional models frequently entail numerous parameters, such as fractional orders and coefficients, which are not directly measurable or specifiable \cite{Jin_AD_2015}. Addressing inverse problems involving fractional-order derivatives presents notable challenges concerning stability, uniqueness, and computational cost.

Physics-Informed Neural Networks (PINNs) \cite{Raissi2019} have emerged as powerful computational frameworks, especially for inverse problems. PINNs have been used to solve a vast range of problems in fluid mechanics \cite{Cai2021, Jin2020}, rheology \cite{Mahmoud2022, viscoelasticNet}, and mass transport \cite{Wang2022, Thakur2024}. Despite their broad applicability, most implementations of PINNs primarily address problems involving integer-order derivatives, typically computed via automatic differentiation. However, a growing interest exists in extending PINNs to handle fractional order derivatives, which offer a more nuanced understanding of complex phenomena. One promising approach involves utilizing the finite difference method to compute fractional order derivatives, thereby paving the way for the development of fractional-PINNs \cite{Pang2019}. This novel methodology has already yielded promising results, facilitating the discovery of fractional order models for turbulent flow \cite{Mehta2019} and epidemiological dynamics \cite{Kharazmi2021}. However, it is worth noting that the L1 scheme \cite{Jin2015}, commonly employed to calculate time-fractional derivatives in these contexts, is tailored exclusively for sub-diffusion with derivatives within the range $(0,1)$, limiting its applicability to capturing only integer-ordered time derivatives.

This study focuses on time-fractional diffusion equations characterizing anomalous diffusion and the fractional Maxwell model governing viscoelastic behavior. To tackle these complex phenomena, we employ a finite-difference method \cite{Lin2007} tailored for fractional orders within the range $[0,1]$. This approach facilitates the calculation of fractional-order derivatives, of which integer-order derivatives are special cases. An overview of fractional derivatives, anomalous diffusion, and fractional viscoelasticity is provided in Sections \ref{frac_derivatives}, \ref{anomalous_diffusion}, and \ref{frac_viscoelasticity}, respectively. Subsequently, in section \ref{PINNs_setup}, we elucidate the methodology devised for solving inverse problems associated with the fractional diffusion equation and the fractional Maxwell model. Notably, our approach ensures that all losses are weighted according to the inherent scales present in the data. Lastly, the outcomes of our investigations into inverse problems are meticulously presented and analyzed in Section \ref{Results}. Through this structured approach, we aim to offer insights into the intricate interplay between fractional calculus and real-world phenomena, shedding light on their underlying mechanisms and behaviors.

\section{Problem setup and methodology}

\subsection{Fractional derivatives} \label{frac_derivatives}

While offering profound insights into complex phenomena, fractional calculus introduces notable computational and analytical challenges. The non-local characteristics of fractional derivatives render them computationally intensive and more challenging to implement than their integer-order counterparts. Moreover, navigating the realm of fractional calculus demands a firm grasp of intricate mathematical concepts, including fractional order operators, differential equations, calculus of variations, and fractional processes. Thus, a robust mathematical foundation coupled with a nuanced understanding of its applications is imperative.
The Caputo fractional derivative of order $\alpha$, for $0 < \alpha < 
 1$ for a function $f$ and time $t$, is defined as \cite{Podlunby1998}
\begin{equation}
  \frac{\partial^\alpha f(t)}{\partial t^\alpha}=\frac{1}{\Gamma(1-\alpha)} \int_0^t \frac{\partial f(s)}{\partial s} \frac{ds}{(t-s)^\alpha},\label{Caputo}
\end{equation}
where $\Gamma$ is the gamma function. For $\alpha =0$, Eq. \eqref{Caputo} corresponds to the classical Helmholtz elliptic equation, and for the limit $\alpha \rightarrow 1$, the integer order time derivative can be obtained. For time-fractional differential equations, it is possible to numerically approximate the value of the fractional derivative. For $ 0 \leq \alpha \leq 1 $, the finite difference approximation of the Caputo fractional derivative ($L_t^\alpha$) of order $\alpha$ can be evaluated as\cite{Lin2007}
\begin{equation}
    L_t^\alpha f\left(t_{k+1}\right):=\frac{1}{\Gamma(2-\alpha)} \sum_{j=0}^k b_j \frac{f(t_{k+1-j}) - f(t_{k-j}) }{\Delta t^{\alpha}},
\end{equation}
where k is the time step number and $\Delta t$ is the time step size. The coefficients $b_j$ are defined as 
\begin{equation}
    b_j := (j+1)^{1-\alpha} - j^{1-\alpha}.
\end{equation}
We rewrite this equation to get
\begin{equation}\label{FD}
    L_t^\alpha f\left(t_{k+1}\right):=\frac{1}{\Delta t^{\alpha} \Gamma(2-\alpha)}  ( b_0 f(t_{k+1}) - b_0 f(t_{k}) +  \sum_{j=0}^{k-1} b_{j+1} f(t_{k-j}) -  \sum_{j=1}^{k} b_{j} f(t_{k-j})).
\end{equation}
\subsection{Anomalous diffusion} \label{anomalous_diffusion}

Diffusion, driven by the erratic thermal motion of molecules, underpins numerous natural phenomena across physical, chemical, and biological realms. In classical Fickian diffusion, transport dynamics align neatly with Gaussian statistics, as evidenced by the linear time evolution of mean squared displacement ($<x^{2}(t)>$). This linear relationship, captured by the equation \cite{Metzler2000TheApproach}:
\begin{equation}
    <x^{2}(t)> \sim D_{1}t,
\end{equation}
where $D_{1}$ is the diffusion coefficient illustrates that the average distance traversed by diffusing particles grows proportionally with the square root of time.

However, in intricate systems like porous media, amorphous semiconductors, and polymeric structures, transport behavior often diverges from this simplistic pattern, heralding what is known as anomalous diffusion. Anomalous diffusion is typified by a non-linear augmentation of mean squared displacement over time, characterized by the power-law expression:
\begin{equation}
    <x^{2}(t)> \sim \Tilde{D}t^{\alpha}.
\end{equation}
Here, $\alpha$ denotes the anomalous diffusion exponent, and $\Tilde{D}$ denotes the generalized diffusion coefficient. The exponent $\alpha$ measures the degree of deviation from normal diffusion. When $\alpha=1$, the diffusion is normal and follows Fick’s law. When $\alpha>1$, the diffusion is faster than Fickian diffusion and is called super-diffusion. When $\alpha<1$, the diffusion is slower than Fickian diffusion and is called sub-diffusion. Typical diffusion processes can be described by the diffusion equation with integer order derivatives 
\begin{equation}
    \frac{\partial c}{\partial t} = \nabla \cdot(D_1 \nabla c),
\end{equation} 
Here, c denotes the concentration of diffusing particles and $\nabla$ denotes the gradient operator. Processes with anomalous diffusion are described by fractional order derivatives. For example, sub-diffusion can be modeled by using a time-fractional derivative of order $\alpha$, where $0<\alpha<1$, as follows
\begin{equation}\label{FDE}
\frac{\partial^{\alpha} c}{\partial t^{\alpha}} = \nabla \cdot(\Tilde{D} \nabla c).
\end{equation}
where $\Tilde{D}$ denotes the sub-diffusion coefficient. This equation implies that the diffusing particles have a long-tailed waiting time distribution, meaning they tend to stay longer in some regions before moving to another. The jump length distribution, however, remains finite. For a concentration-dependent diffusion coefficient, we can write \cite{Adib_2022}
\begin{equation} \label{ADE}
    \frac{\partial^{\alpha} c}{\partial t^{\alpha}} =  \Tilde{D} (c_{xx} + c_{yy}) + \Tilde{D}_c(c_x^2 + c_y^2),
\end{equation}
where the subscripts $x$, $y$ and $c$ denote the derivatives with respect to the $x$-coordinate, $y$-coordinate and the concentration, respectively.  

\subsection{Fractional viscoelasticity}
\label{frac_viscoelasticity}
The theory of linear viscoelasticity offers a robust mathematical framework for understanding the interplay between stress ($\tau(t)$), strain ($\epsilon (t)$), and time ($t$). A common examination method, known as a relaxation test, involves subjecting a material to a constant strain and analyzing its stress response. In such tests, the resulting stress ($\tau(t)$) can be expressed as:
\begin{equation}
    \tau(t) = G(t)\epsilon^0,
\end{equation}
where $G(t)$ is defined as the relaxation modulus and $\epsilon^0$ is the strain applied at the initial time. The traditional linear visoelastic models are built using springs and dashpots. An ideal spring, obeying Hooke's law, relates stress and strain as: 
\begin{equation}
    \tau(t)  = E \epsilon(t),
\end{equation}
while a dashpot, adhering to Newton's model, equates stress to the derivative of strain with respect to time: 
\begin{equation}
    \tau(t)  = \eta \frac{d\epsilon(t)}{dt},
\end{equation}
where E is the elastic modulus and $\eta$ is the material viscosity. These elements can be interconnected to form more complex models, with the relaxation modulus expressions derived accordingly. Connecting a spring and a dashpot in series yields the Maxwell model, while parallel connection results in the Kelvin-Voigt model. Introducing additional elements facilitates the capture of multiple timescales in material behavior.
However, the incorporation of numerous elements poses challenges. Interpretation becomes more intricate, and the model becomes increasingly susceptible to overfitting as the number of elements rises. An alternative approach to capturing complex material behavior with fewer parameters involves the utilization of a viscoelastic element known as the springpot. In the case of a springpot, the stress-strain relationship is defined as:
\begin{equation}
    \tau (t) = K_{\alpha}\frac{d^{\alpha} \epsilon(t)}{dt^{\alpha}},
\end{equation}
where $K_{\alpha}$ is the `firmness' of the material and $\alpha$ is the order of the fractional derivative. The springpot's behavior varies with the value of the fractional derivative order. The stress-strain relationship for a fractional Maxwell model is given by \cite{Mainardi2010}
\begin{equation}\label{FracMaxwell}
        \tau (t) + \eta\frac{d^{\nu} \tau(t)}{dt^{\nu}} = \kappa\frac{d^{\nu} \epsilon(t)}{dt^{\nu}},
\end{equation} 
with the relaxation modulus being defined as
\begin{equation}\label{Rel_mod}
    G(t) = \frac{\kappa}{\eta} E_{\nu}\left(-\frac{t^{\nu}}{\eta}\right),
\end{equation}
where $E_{\nu}$ is Mittag-Leffler function of order $\nu$. The Mittag-Leffler function is defined as
\begin{equation}
    E_{\nu}(z)=\sum_{n=0}^{\infty} \frac{z^n}{\Gamma(\nu n+1)},
\end{equation}
where $\Gamma$ is the gamma function. The fractional Maxwell model has the capability of capturing long memory effects and can capture complex material behavior with fewer parameters.

\subsection{Physics-Informed Neural Networks}\label{PINNs_setup}
\begin{figure}
    \centering
    \includegraphics[width=\textwidth]{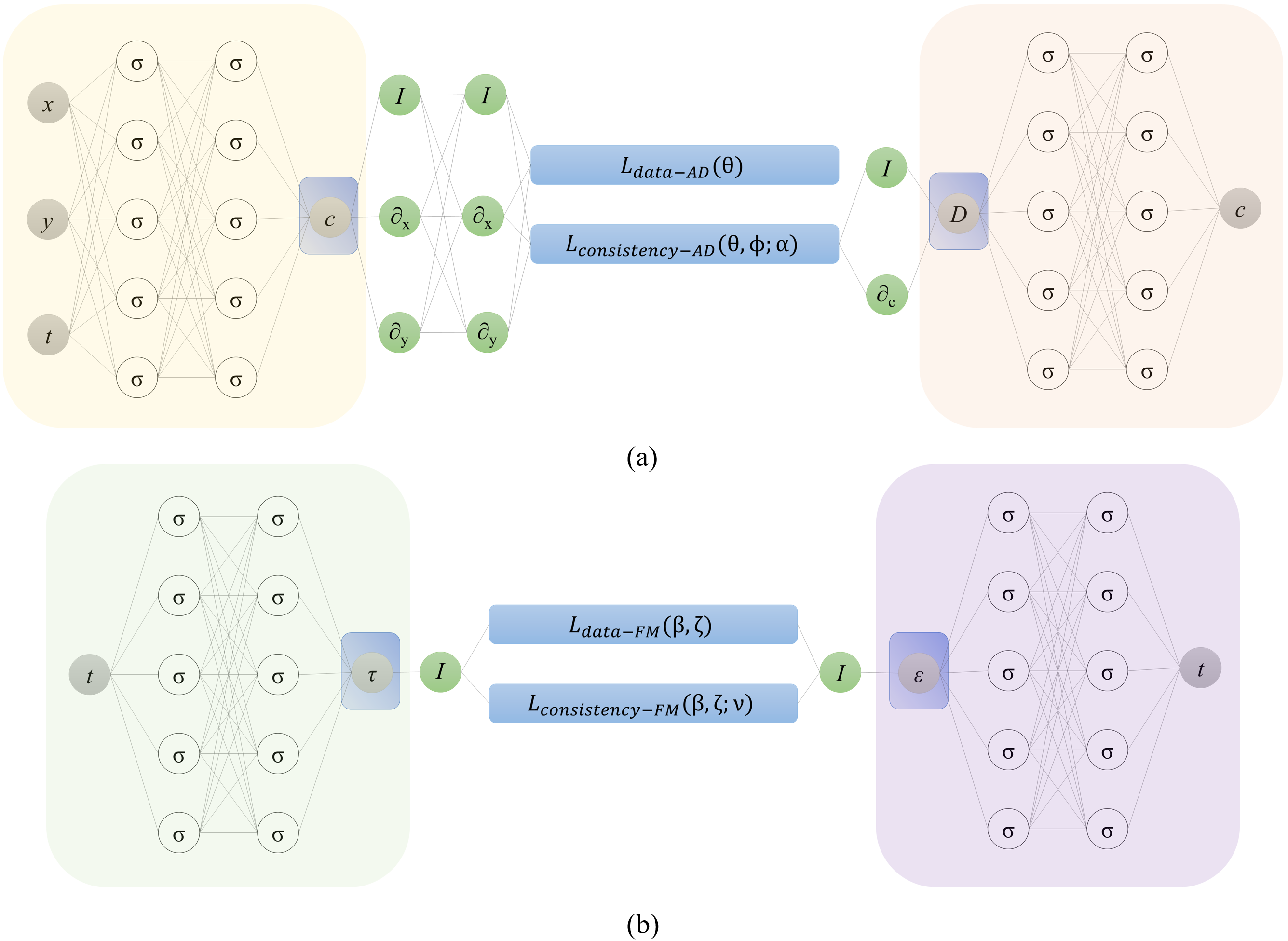}
    \caption{The problem set up for (a) anomalous diffusion and (b) the fractional Maxwell model, as described in section \ref{PINNs_setup}. Here, I denotes the identity operator, and the differential operators $\partial x$, $\partial y$ and $\partial c$ are computed using automatic differentiation. }
    \label{fig:NN}
\end{figure}
While automatic differentiation enables the computation of integer-order derivatives, direct calculation of fractional-order derivatives presents a challenge. To address this, in computing the residual of a time-fractional differential equation, we employ automatic differentiation for spatial order derivatives and finite differences for time derivatives \cite{Pang2019}, utilizing equation \eqref{FD}. 

In our pursuit to infer the anomalous diffusion coefficient as a function of concentration and fractional order derivative, we approximate the functions $(t,x,y) \mapsto (c^{pu})$ and $(c^{pu}) \mapsto (D)$ using two neural networks parameterized by $\theta$ and $\phi$, respectively. Here, the superscripts `$pu$' denote the physics uninformed nature of these networks.

To quantify the loss for regression over the concentration field, we define the mean squared loss as follows:
\begin{equation}
    L_{data - AD}(\theta) =  \mathbb{E}_{(t,x,y,c)}[\frac{|{c^{pu}}(t,x,y;\theta) - {c}|^2}{{\sigma_c}^2}],
\end{equation}
where $\mathbb{E}$ denotes the expectation approximated by the population mean, ${\sigma_c}$ is the standard deviation of the reference concentration field $c$ and the subscript $AD$ is used to denote anomalous diffusion. We now define 
\begin{equation}
    h(c,t^k) = b_0 c(t^k) - \sum_{j=0}^{k-1} b_{j+1} c(t^{k-j}) + \sum_{j=1}^{k} b_{j} c(t^{k-j}),
\end{equation}
where $t^k$ is the $k^{th}$ time step. Further, let
\begin{equation}
    g(c,D) = {D} (c_{xx} + c_{yy}) + {D}_c((c_x)^2 + (c_y)^2),
\end{equation}
as defined in Eq. \eqref{ADE}.
To construct the physics-informed network for anomalous diffusion, we combine the equations \eqref{FD} and \eqref{ADE} to get
\begin{equation}
    c^{pi}(t^{k+1},\boldsymbol{x};\Delta t, \theta, \phi, \alpha) = \Gamma(2-\alpha) \Delta t^{\alpha}g(c^{pu}(t^k,\boldsymbol{x}; \theta), \Tilde{D}(c^{pu}; \phi)) + h(c^{pu}(t^k,\boldsymbol{x}; \theta),t^k),
\end{equation}
where $\boldsymbol{x} = (x,y)$ are the spatial coordinates and the superscript `$pi$' is used to denote `physics-informed'. Since the physics-uninfomed and physics-informed concentration fields evaluated at the same spatio-temporal point need to be consistent, this allows us to define a consistency loss \cite{Thakur2023} as 
\begin{equation}
    L_{consistency-AD}(\theta, \phi; \Delta t, \alpha) =  \mathbb{E}_{(t,\boldsymbol{x})}[\frac{|{c}^{pi}(t,\boldsymbol{x};\Delta t,\theta, \phi, \alpha)-{c}^{pu}(t,\boldsymbol{x};\theta)|^2}{{\sigma_{c}}^2}].
\end{equation}
The parameters $\theta$ and $\phi$ can now be optimized by minimizing the following mean squared error function
\begin{equation} \label{MSE_AD}
    L_{MSE-AD}(\theta, \phi; \alpha) = L_{data - AD}(\theta) +  L_{consistency-AD}(\theta, \phi; \alpha).
\end{equation}
Similarly, to learn the parameters of the fractional Maxwell equation (Eq. \eqref{FracMaxwell}), we utilize data on stress ($\tau$) and strain ($\epsilon$) as functions of time ($t$). We approximate the functions $ (t) \mapsto (\sigma^{pu})$ and $(t) \mapsto (\epsilon^{pu})$ using two fully-connected neural networks parameterized by $\beta$ and $\zeta$, respectively. The mean squared loss for regression over the stress and strain fields is defined as:
\begin{equation}
    L_{data - FM}(\beta, \zeta) =  \mathbb{E}_{(t,\tau)}[\frac{|{\tau^{pu}}(t;\beta) - {\tau}|^2}{{\sigma_{\tau}}^2}] + \mathbb{E}_{(t,\epsilon)}[\frac{|{\epsilon^{pu}}(t;\zeta) - {\epsilon}|^2}{{\sigma_{\epsilon}}^2}],
\end{equation}
where ${\sigma_{\tau}}$ and ${\sigma_{\epsilon}}$ are the standard deviations of the reference stress field $\tau$ and the reference strain field, respectively. To get the physics-informed stress, we combine equations \eqref{FD} and \eqref{FracMaxwell} to derive
\begin{equation}
    \tau^{pi}(t^{k+1};\Delta t, \beta, \zeta, \nu) = \kappa L_t^\nu (\epsilon^{pu}(t^{k+1};\Delta t, \zeta)) - \eta L_t^\nu (\tau^{pu}(t^{k+1};\Delta t, \beta)),
\end{equation}
which allows us to define
\begin{equation}
    L_{consistency-FM}(\beta, \zeta; \Delta t, \nu) =  \mathbb{E}_{(t)}[\frac{|{\tau}^{pi}(t;\Delta t,\beta, \zeta, \nu)-{\tau}^{pu}(t;\beta)|^2}{{\sigma_{\tau}}^2}],
\end{equation}
where the subscript $FM$ is used to denote fractional Maxwell. The parameters $\beta$ and $\zeta$ are then optimized by minimizing the following mean squared error function
\begin{equation} \label{MSE_FM}
    L_{MSE-FM}(\beta, \zeta) = L_{data - FM}(\beta, \zeta) +  L_{consistency-FM}(\beta, \zeta; \nu).
\end{equation}
\section{Results and discussion} \label{Results}
\subsection{Numerical dataset - anomalous diffusion}
We constructed a synthetic dataset to serve as a benchmark and evaluate the efficacy of our framework. Our neural network architecture comprised a fully connected network with 4 layers and 16 neurons per layer to approximate the function mapping $(t,x,y) \rightarrow (c^{pu})$, and another fully connected network with 2 layers and 4 neurons per layer to approximate the subsequent mapping $(c^{pu}) \rightarrow (D)$. Throughout this study, we employed the swish activation function for all neural networks.

In our training process, we adopted a cosine learning rate schedule \cite{Ilya2017}. Specifically, we set the maximum learning rate ($\eta_{max}$) to $2.5\text{e-}03$ and the minimum learning rate ($\eta_{min}$) to $2.5\text{e-}06$, enabling us to compute the learning rate ($\eta$) dynamically using the following formula:

\[
A =  A_{min} + 0.5(A_{max} - A_{min})\left( 1 + \cos\left( \frac{T_{cur}}{T_{max}}\pi\right)\right),
\]
where $A_{min}$ and $A_{max}$ are the minimum and maximum values of the learning rate, respectively, $T_{cur}$ denotes the current time step and $T_{max}$ represents the total number of time steps. The neural networks underwent training for 100,000 iterations and the optimized parameters were obtained by minimizing the loss in Eq. \eqref{MSE_AD}.

\begin{figure}
    \centering
\includegraphics[width=0.6\textwidth]{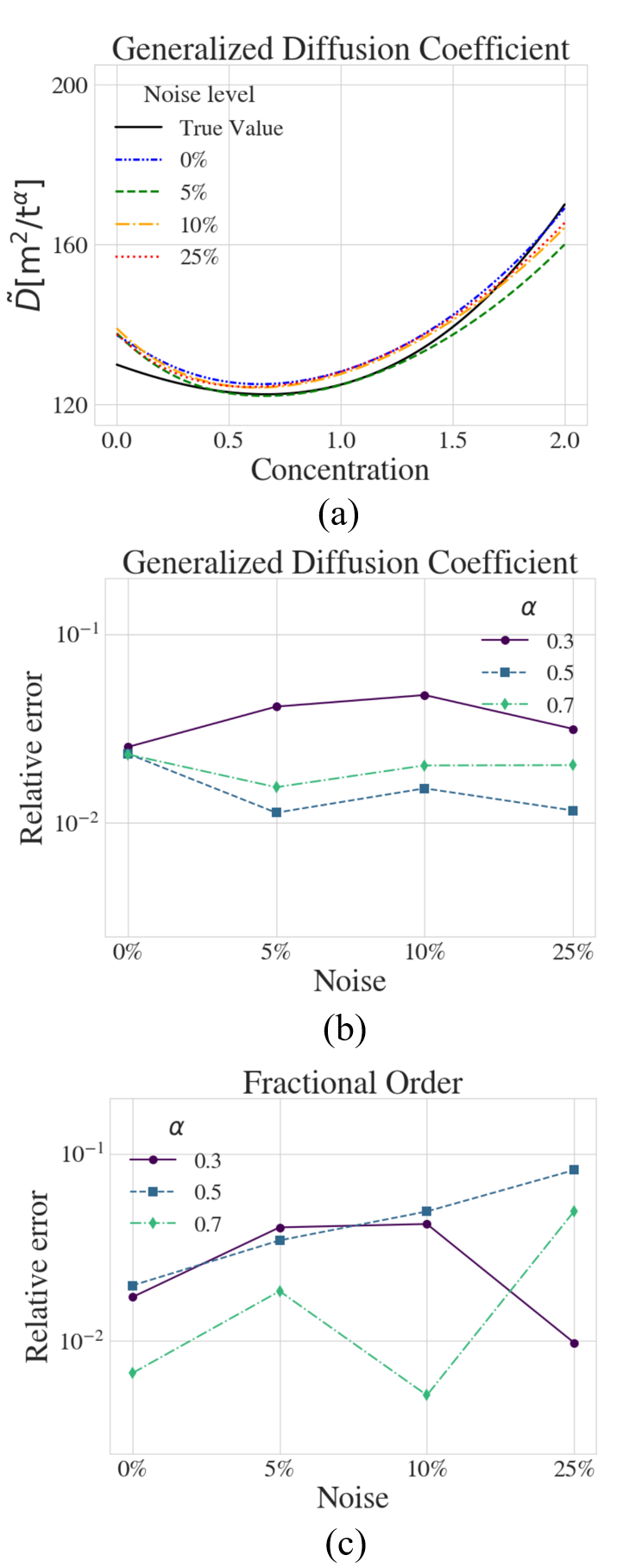}
    \caption{(a) The reference generalized diffusion coefficient compared to different levels of added noise. The relative errors in the (b) generalized diffusion coefficient and (c) fractional order for different levels of added noise.}
    \label{fig:result_AD}
\end{figure}

To assess the robustness of our framework, we introduced Gaussian noise to the concentration field. The resulting predicted generalized diffusion coefficients are visualized in Fig. \ref{fig:result_AD}, along with the relative errors in the predicted generalized diffusion coefficient and fractional derivative order. Our solver effectively captured both the generalized diffusion coefficient and the fractional derivative order across all tested scenarios.

\subsection{Experimental dataset - fractional Maxwell model}

\begin{figure}
    \centering\includegraphics[width=\textwidth]{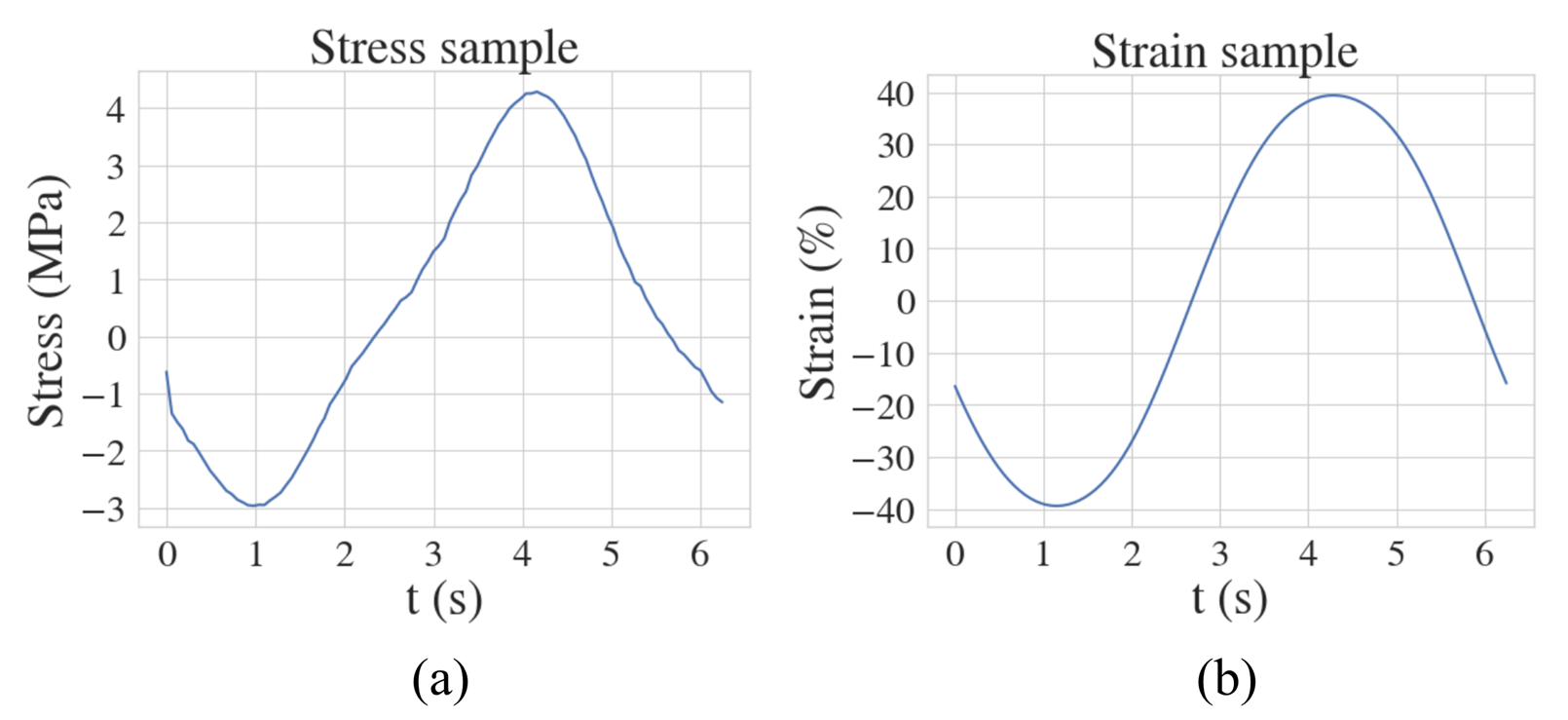}
    \caption{Plot of the experimentally obtained (a) stress and (b) strain values for a sample reported in \cite{mitra2024rheological}. }
    \label{fig:sample}
\end{figure}

To evaluate our framework's performance on experimental data to learn the parameters of the fractional Maxwell model, we leverage the nonlinear stress-strain and corresponding relaxation modulus data obtained from shear rheology experiments conducted on minipig tissues by Mitra et al. \cite{mitra2024rheological}. We plot a sample of the experimentally obtained stress and strain curves in Fig. \ref{fig:sample}. Our neural network architecture consists of a fully connected network with 2 layers and 20 neurons per layer, tasked with approximating the function mapping $(t) \rightarrow (\sigma^{pu})$. Additionally, another fully connected network with the same architecture is employed to approximate the subsequent mapping $(t) \rightarrow (\epsilon^{pu})$.

\begin{figure}
    \centering
    \includegraphics[width=0.85\textwidth]{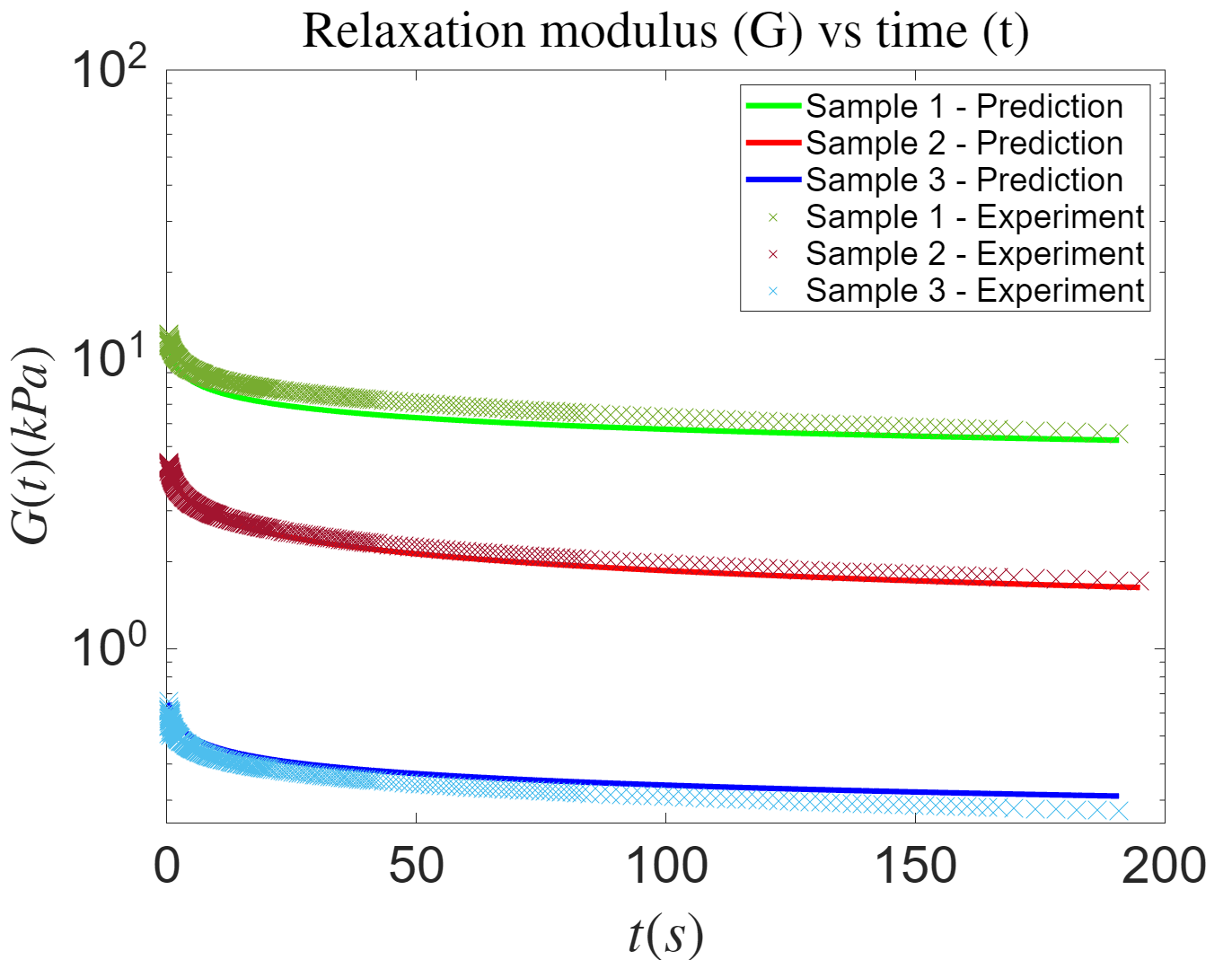}
    \caption{The predicted relaxation modulus and the experimental data corresponding to three samples of minipig skin tissue. }
    \label{fig:result_FM}
\end{figure}

We iteratively learn the parameters of equation \eqref{FracMaxwell} by minimizing the loss function defined in equation \eqref{MSE_FM}. Subsequently, we utilize these learned parameters to predict the relaxation modulus as defined in equation \eqref{Rel_mod}. Our study encompasses three distinct tissue samples: neck, belly, and breast tissue. The outcomes of our analysis are visualized in Fig. \ref{fig:result_FM}. Additionally, we provide detailed numerical values for the relative errors in Table \ref{ta:rel_err_FM}. Notably, all relative errors for the predicted relaxation moduli concerning the reported values are found to be less than 10\%.

\begin{table}[ht]\label{Result_Summary_FM}
  \caption{The relative errors for the relaxation moduli (G(t)) for the three pig tissue samples.}
  \begin{center}
    \begin{tabular}{|c|c|c|c|}
      \hline
      \bf Samples & \bf Sample 1& \bf Sample 2& \bf Sample 3\\
      \hline
      Relative error & $9.14\times10^{-2}$ & $3.06\times10^{-2}$& $8.92\times10^{-2}$\\
      \hline
      
    \end{tabular}
  \end{center}
  \label{ta:rel_err_FM}
\end{table}
\section{Conclusions and future work}

Time-fractional differential equations find widespread applications across various fields, albeit often posing computational challenges and instability issues, especially in the context of inverse problems. Recognizing this, there is growing interest in leveraging Physics-Informed neural network-based frameworks to address inverse problems entailing time-fractional derivatives. In such frameworks, the fractional time derivative can be efficiently computed using finite differences, while automatic differentiation handles other derivatives.

In this study, we tackle two inverse problems associated with 1) anomalous diffusion and 2) fractional viscoelasticity employing the aforementioned approach. Our methodology involves defining the residual loss in a manner that facilitates scaling the loss terms with the standard deviation of the observed data. We utilize numerically generated datasets and experimental data for learning the fractional coefficient and the concentration-dependent generalized diffusion coefficient, as well as for calibrating parameters for the fractional Maxwell model using stress and strain data over time.

To evaluate the performance of our framework in handling anomalous diffusion, we conducted rigorous testing by comparing our predictions for a concentration-dependent generalized diffusion coefficient and the fractional order of the derivative against synthetically generated reference values. Remarkably, even with the addition of a 25\% Gaussian noise to the concentration dataset, our framework demonstrated robust performance. Specifically, we observed that the relative error in predicting the generalized diffusion coefficient and the order of the fractional derivative was less than 10\%. This outcome underscores the resilience and accuracy of our framework, even in the presence of significant noise, reaffirming its reliability in practical applications.

To validate our findings, we predict relaxation moduli for three distinct samples of pig tissue and compare them against reported values in the literature. The relative errors are consistently below 10\%, highlighting the efficacy of the fractional model which requires fewer parameters. Furthermore, our approach holds promise for extending to model non-linear fractional viscoelasticity and can readily incorporate experimental data to validate anomalous diffusion predictions. Additionally, the solver's applicability can be extended to three-dimensional scenarios, offering a broader scope for future exploration into equations involving time-fractional derivatives.
\bibliographystyle{unsrt}
\bibliography{references}
\end{document}